\def\year{2022}\relax
\documentclass[letterpaper]{article} 
\usepackage{aaai22}  
\usepackage{times}  
\usepackage{helvet}  
\usepackage{courier}  
\usepackage[hyphens]{url}  
\usepackage{graphicx} 
\usepackage{natbib}  
\usepackage{caption} 
\DeclareCaptionStyle{ruled}{labelfont=normalfont,labelsep=colon,strut=off} 
\usepackage{algorithm}
\usepackage{algorithmic}

\usepackage{newfloat}
\usepackage{listings}
\floatstyle{ruled}
\newfloat{listing}{tb}{lst}{}
\floatname{listing}{Listing}

\makeatletter
\newcommand{\Ni}{({\em i})~}
\newcommand{\Nii}{({\em ii})~}
\newcommand{\Niii}{({\em iii})~}
\newcommand{\Niv}{({\em iv})~}
\usepackage{amssymb}
\usepackage{amsmath}
\usepackage{pifont}
\def\eg{\emph{e.g}.}

\usepackage{xcolor}
\newcommand{\prj}[1]{\textcolor{black}{#1}} 
\newcommand{\prjj}[1]{\textcolor{black}{#1}} 
\newcommand{\jhh}[1]{\textcolor{black}{#1}} 
\newcommand{\jh}[1]{\textcolor{black}{#1}} 
\newcommand{\sh}[1]{\textcolor{black}{#1}} 
\usepackage{multirow}
\usepackage{array}
\makeatother

\title{AutoBERT-Zero: Evolving BERT Backbone from Scratch}
\author {
    Jiahui Gao\textsuperscript{\rm 1},
    Hang Xu\textsuperscript{\rm 2}\thanks{Corresponding author.},
    Han Shi\textsuperscript{\rm 3},
    Xiaozhe Ren\textsuperscript{\rm 2}, \\
    Philip L.H. Yu\textsuperscript{\rm 4},
    Xiaodan Liang\textsuperscript{\rm 5},
    Xin Jiang\textsuperscript{\rm 2},
    Zhenguo Li\textsuperscript{\rm 2}
}
\affiliations {
    \textsuperscript{\rm 1} The University of Hong Kong,
    \textsuperscript{\rm 2} Huawei Noah’s Ark Lab\\
    \textsuperscript{\rm 3} Hong Kong University of Science and Technology\\
    \textsuperscript{\rm 4} The Education University of Hong Kong, 
    \textsuperscript{\rm 5} Sun Yat-sen University, China\\
}

\begin{document}

\maketitle

\begin{abstract}
Transformer-based pre-trained language models like BERT and its variants have recently achieved promising performance in various natural language processing (NLP) tasks. However, the conventional paradigm constructs the backbone by purely stacking the manually designed global self-attention layers, introducing inductive bias and thus leads to sub-optimal. In this work, \prjj{we make the first attempt to automatically discover novel pre-trained language model (PLM) backbone 
on a flexible search space containing the most fundamental operations from scratch.
Specifically, we propose a}
well-designed search space which \Ni contains primitive math operations in the intra-layer level to explore novel attention structures, and \Nii leverages convolution blocks to be the supplementary for attentions in the inter-layer level to better learn local dependency.
\prjj{To enhance the efficiency for finding promising architectures, we propose an Operation-Priority Neural Architecture Search (OP-NAS) algorithm, which}
optimizes both the search algorithm and evaluation of candidate models.
Specifically, we propose Operation-Priority (OP) evolution strategy to facilitate model search via balancing exploration and exploitation.
Furthermore, we design a Bi-branch Weight-Sharing (BIWS) training strategy for fast model evaluation.
Extensive experiments show that the searched architecture (named \textbf{AutoBERT-Zero}) significantly outperforms BERT and its variants of different model capacities in various downstream tasks, proving the architecture's transfer and scaling abilities. Remarkably, AutoBERT-Zero-base outperforms RoBERTa-base (using much more data) and BERT-large (with much larger model size) by 2.4 and 1.4 higher score on GLUE test set.
\end{abstract}


\section{Introduction} \label{sec:intro}
Benefiting from the powerful capacity of self-attention structures in transformers \citep{vaswani2017attention}, the pre-trained language models (PLM) (\eg\ BERT \citep{devlin2019bert}, \ RoBERTa \citep{liu2019roberta}, ALBERT \citep{lan2019albert}, GPT3 \citep{brown2020language}) have achieved satisfying performance across various NLP tasks \citep{wang2018glue,rajpurkar-etal-2016-squad,rajpurkar2018know,zellers-etal-2018-swag}. 
\jhh{All these models are based on the fixed hand-crafted self-attention structure 
by varying training resources, parameter numbers, layer numbers and inputs. }

The conventional paradigm constructs the backbone by stacking the manually-designed global self-attention layers. However, many recent works have pointed out that the design of self-attention structures is not optimal
(Kovaleva et al. \citeyear{kovaleva2019revealing}; Michel et al. \citeyear{michel2019sixteen}; Dong et al. \citeyear{dong2021attentionnot}), whose inductive bias limits its performance as well as efficiency. 
In particular, (Dong et al. \citeyear{dong2021attentionnot}) find that repeatedly stacking self-attention results to ``token-uniformity'' problem, meaning that different tokens are mapped to similar latent representations. Even though they claim that skip connection and multi-layer perceptions mitigate this  problem, we still observe it on the BERT output (see Figure~\ref{fig:cosine_rank}). 
Another work Reformer (Kitaev et al. \citeyear{kitaev2020reformer}) discovered that sharing the weights for query and key does not impact the model’s performance, indicating that redundant parameters exist in self-attention structure. In addition, ConvBERT \citep{jiang2020convbert} shows that local operations such as convolution helps better learn the inherent local dependencies in natural languages. Here, we raise the following questions: Does there exist more powerful and efficient attention beyond the pure query-key-value self-attention for PLM? Can we boost the model performance and efficiency by flexibly combining global attention with local operations?

\begin{figure}[t]
\centering
	\includegraphics[width=0.45\textwidth]{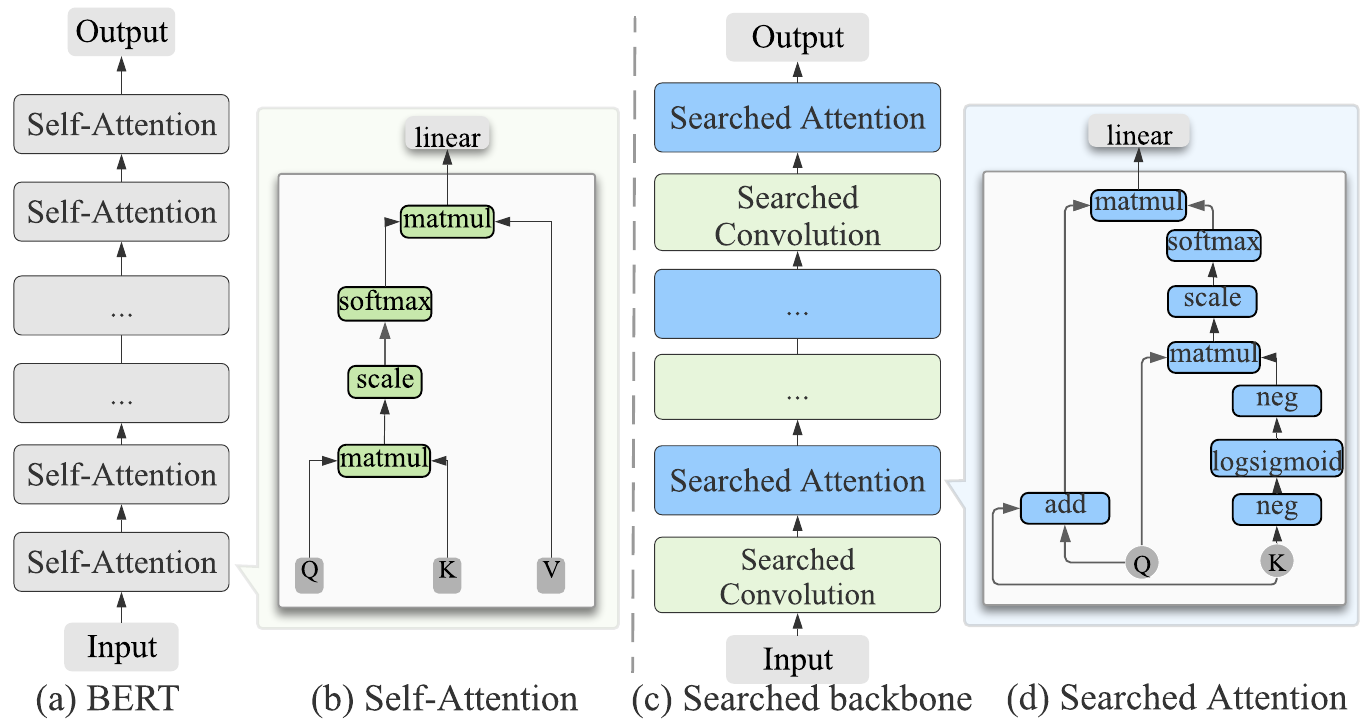}
	\caption{\small{Comparison between BERT and our searched model. Our searched AutoBERT-Zero is a hybrid structure with convolution layers and the novel searched attention layers, whose kernel sizes and attention structures are various across different layers.}}
	\label{fig:archs_compare}
	\vspace{-8mm}
\end{figure}
To address the above fundamental challenges in the NLP field, we resort to Neural Architecture Search (NAS), which has emerged as a powerful technique to automatically discover promising models without excessive human intervention and tedious tuning. NAS is empowered by a search algorithm and a well-designed search space. 
The effectiveness of NAS is validated on many computer vision tasks (e.g., image classification \citep{zoph2016neural, shi2020bridging},
object detection 
\citep{xu2019auto,yao2021jointdetnas}
. Nevertheless, few works leverage NAS to design backbone structure for PLM. The only related works, AdaBERT \citep{2020-adabert} and DynaBERT \citep{DBLP:conf/NeurIPS/HouHSJCL20} use NAS to compress the full-sized BERT into small models, \jh{while Evolved Transformer \citep{so2019evolved} searches architecture on specific downstream tasks.
Besides, as architectures in AdaBERT and Evolved Transformer are task-specific, those models are not applicable for general NLP tasks. Meanwhile, the searched models in DynaBERT and Evolved Transformer are still transformer-based, which does not explore more powerful attention structure.}

To the best of our knowledge, using NAS to discover a novel general PLM backbone from scratch has not been investigated.
\prjj{In this work, we aim to explore powerful PLM backbone by discovering novel attention structures as well as whole backbone architecture from a flexible search space.} Specifically, we design both intra-layer and inter-layer search spaces \prjj{that provide a wide variety of candidate architectures to} prevent the inductive bias in conventional transformer. The intra-layer search space \prjj{with few constraints enables finding novel self-attention mechanism, which contains various primitive mathematical operations to construct computation graph with variable path length and flexible input nodes.} 
The inter-layer search space contains global (self-attention) and local operations (convolution) on the backbone level, which provides flexibility in learning global and local dependencies at different layers.

Since pretraining a PLM is quite time consuming, the computational burden of NAS for PLM is much more overwhelming than utilizing NAS for CV tasks, \prjj{especially given that our search space is extremely huge}.
\prj{Thus, it is crucial to make the NAS algorithm more efficient \prjj{in terms of both speed and memory}. 
\prjj{To this end, we propose a novel Operation-Priority Neural Architecture Search (OP-NAS) algorithm.
During search phase}, we promote Operation-Priority (OP) evolution strategy. This strategy \prjj{leverages prior information of operations at each position in the computation path to flexibly balance exploration and exploitation when mutating new architectures, which escapes local optimal and speeds up the search.}
To facilitate model evaluation, we design Bi-branch Weight-Sharing (BIWS) training strategy, which introduces a super-net to keep track of the trained weights for both the attention structures and convolution blocks \jh{on each layer}. The candidates are initialized with the weights extracted from the super-net during evaluation to prevent repeated pretraining.}

Extensive experiments are conducted on the widely used Natural Language Understanding(NLU) and Question Answering(QA) benchmarks.
The best searched architecture(named \textbf{AutoBERT-Zero}) is shown on Figure~\ref{fig:archs_compare}(c), which stacks novel searched attention structures and convolutions. Our AutoBERT-Zero achieves 87.7 GLUE score when trained on the commonly used vallina pre-train tasks, consistently outperforming current state-of-the-art (SOTA) methods by a large margin (4.1 higher than T5), while requiring fewer parameters (52.7\% fewer parameters than T5).
More remarkably, our AutoBERT-Zero-base surpasses RoBERTa-base (using much more data) and BERT-large (with much larger model size) by 2.4 and 1.4 higher score on GLUE test set.

Our main contributions are summarized as follows:
\Ni This is the first work conducting NAS to automatically discover new self-attention structures and better backbones for PLM. 
\Nii The well-designed search space allows flexible variations in self-attention structures/input nodes/combinations of local and global operations, which enables deriving powerful architectures. 
\Niii The proposed OP evolution algorithm and BIWS training significantly accelerate the model search and evaluation.
\Niv Extensive downstream evaluations demonstrate the effectiveness and scaling ability of the searched model AutoBERT-Zero.


\vspace{-2mm}
\section{Related Works}\label{sec:related_work}
\textbf{Pre-trained Language Model (PLM).}
Recently, the transformer-like paradigm \citep{vaswani2017attention,radford2018improving} has dominated the research on pre-trained language models.
BERT \citep{devlin2019bert} achieves SOTA performance in various NLU tasks by stacking the encoder of the transformer.
Later, diverse BERT variants appear. For example, UniLM~\citep{dong2019unified}, XLNet~\cite{yang2019xlnet}, ELECTRA~\cite{clark2019electra} introduce new pre-training objectives;
Synthesizer \citep{tay2020synthesizer} considers using random matrices to replace the dot-product self-attention mechanism; ConvBERT \citep{jiang2020convbert} replaces part of attention heads with span-based convolution. However, to the best of our knowledge, apart from ConvBERT and Synthesizer, no other work challenges the transformer-based backbone that purely uses the dot-product self-attention module.
In this work, we delve into a more general formulation of attention expression by the combination of primitive math operations.

\textbf{Neural Architecture Search (NAS).} 
Early NAS methods search SOTA architectures based on reinforcement learning \citep{zoph2016neural}, which is computationally expensive. Subsequently, AmoebaNet \citep{real2019regularized} applies the evolution algorithm for NAS. \prjj{More EA-based methods were further proposed, which exploit the evaluated candidates by modifying how the population list is maintained \cite{zhu2019eena, liu2019deep}.
}
Gradient-based methods such as DARTS \citep{liu2018darts} were designed to speed up the model search at the expense of higher memory consumption. 
More recently, AutoML-Zero \citep{real2020automlzero} proves that using the basic mathematical operators can successfully develop a machine learning algorithm.

\begin{figure*}[t]
\centering{}\vspace{-2mm}
\includegraphics[width=1.0\textwidth]{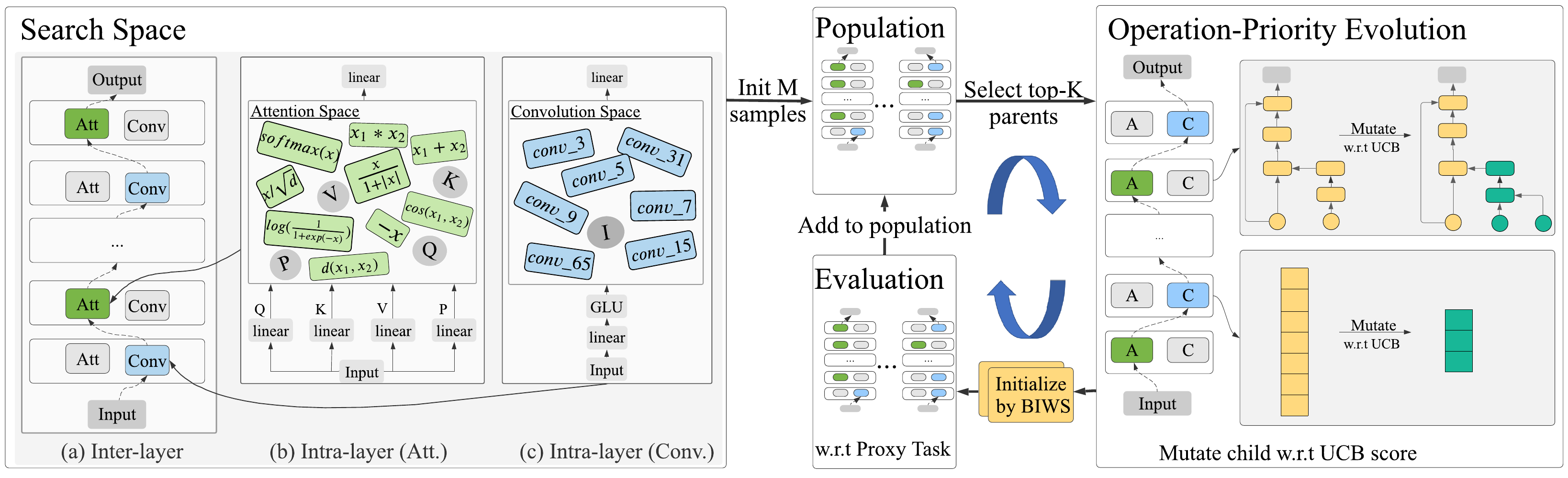} \vspace{-4mm}
\caption{\label{fig:flow_chart}An overview of our OP-NAS framework
for pre-trained language models. Our method directly searches better
backbone architectures from scratch (using primitive operations). We propose
a hierarchical search space for exploring new self-attention structures
and an efficient combination of local and global dependencies. By
\jh{introducing } operation-priority(OP) evolution algorithm with BIWS strategy,
our method efficiently searches over a wide range of the possible
\jh{arichitecures.}
}
\vspace{-5mm}
\end{figure*}

\textbf{NAS for Pre-trained LM.}
Despite the satisfying performance in CV fields, for pre-trained language model, NAS methods are only adopted to BERT compression. AdaBERT \citep{2020-adabert} first introduces NAS to compress BERT into small models using traditional convolution operations. However, the searched architectures are task-specific rather than general pre-trained language models.
DynaBERT \citep{DBLP:conf/NeurIPS/HouHSJCL20} proposes a training method allowing compression in both width and depth directions w.r.t the full-sized teacher BERT model, whose searched models are still 
transformer
backbones.
Orthogonal to the above methods, inspired by the view of AutoML-Zero, we design a search space containing primitive operators and propose a novel NAS method to
\jhh{develop novel attention structure and backbone for general PLM from scratch.
}

\section{Methods}\label{sec:Methods}

In this section, we present an efficient
PLM architecture searching pipeline that evolves the backbone from scratch, as shown in Figure \ref{fig:flow_chart}. We first introduce our hierarchical coarse-to-fine search space, then elaborate on our operation-priority Neural Architecture Search (OP-NAS) algorithm.

\subsection{Search Space Design\label{search_space}}

\prj{We design a two-level search space for discovering novel self-attention structures as well as an overall efficient PLM backbone:}
\Ni \textbf{intra-layer level search space} enables exploring new self-attention structures from primitive operation level; 
\Nii \textbf{inter-layer level search space} leverages global attention layers and local convolution towards an efficient combination of local and global dependencies.

\subsubsection{Intra-layer Search Space}

As shown in Figure \ref{fig:archs_compare}(b), the original self-attention head can be expressed as follows:
\vspace{-1mm}
\begin{small}
\begin{align}
    Attn(X)
    &=\sigma(XW_{Q}(XW_{K})^{\top}/\sqrt{d_h})XW_{V}W_{O}^{\top}\\
    &=\sigma(QK^{\top}/\sqrt{d_h})VW_{O}^{\top}, \label{eq:attn}
\end{align}
\label{equ:self-att}
\end{small}
\hspace{-1.5mm}where $X\in\mathbb{R}^{n\times d}$ is the input, $\sigma$ is softmax
function and self-attention layer is parametered by $W_{Q}^{k},W_{K}^{k},W_{V}^{k},W_{O}^{k}\in\mathbb{R}^{d\times d_{h}}(d_{h}=d/H)$.
The input nodes for a
typical self-attention layer are calculated by three fully connected layers from the inputs, called \textit{query} ($Q=XW_Q$), \textit{key} ($K=XW_K$) and \textit{value} ($V=XW_V$).
We raise two questions: (a) Can we use fewer inputs (e.g., two inputs) to make the transformer more efficient? (b) Can we build a more powerful self-attention architecture by incorporating various mathematical operations?

\textbf{(1) Flexible Input Nodes.}
For question (a), we allow flexible number of
input nodes for our self-attention architecture. More specifically,
we add another input node $P$ to construct a search space with four
input nodes, where $P$ is mapped through another linear transformation
matrix from the original input $(P=XW_P)$. Different from the original transformers
with fixed three input nodes, our intra-layer search space allows a
range of $2\sim 4$ input nodes.

\begin{table}
\resizebox{0.45\textwidth}{!}{
\begin{tabular}{lll}
\hline 
{\small{}Type} & {\small{}Operation} & {\small{}Expression}\tabularnewline
\hline 
\multirow{6}{*}{{\small{}unary}} & {\small{}neg} & {\small{}$-x$}\tabularnewline
 & {\small{}transpose} & {\small{}$x^{\top}$}\tabularnewline
 & {\small{}scale} & {\small{}$x/\sqrt{d_x}$}\tabularnewline
 & {\small{}softmax} & {\small{}$softmax(x)$}\tabularnewline
 & {\small{}logsigmoid} & {\small{}$log({1}/{(1+exp(-x))})$}\tabularnewline
 & {\small{}softsign} & {\small{}${x}/(1+|x|)$}\tabularnewline
\hline 
\multirow{4}{*}{{\small{}binary}} & {\small{}add} & {\small{}$x_{1}+x_{2}$}\tabularnewline
 & {\small{}matmul} & {\small{}$x_{1}\cdot x_{2}$}\tabularnewline
 & {\small{}cosine similarity} & {\small{}$cos(x_{1},x_{2})$}\tabularnewline
 & {\small{}euclidean distance} & {\small{}$d(x_{1},x_{2})$}\tabularnewline
\hline 
\end{tabular}}
\caption{\label{table: operations_attention}Mathematical primitive operations
in our Intra-layer Search Space. We try to find a better self-attention
structure by construct those operations in a DAG computation graph.}
\vspace{-6mm}
\end{table}
\textbf{(2) Primitive Operations.} The key component of transformer architecture
is the self-attention layer, which first generates an attention matrix, then use it to calculate the weighted sum of \textit{values}. The attention
matrix measures the similarity between the \textit{queries} and \textit{keys}.
For question (b), we enable finding a better structure of self-attention by designing a more flexible primitive operation search space. Rather than only using $\langle matmul\rangle$ and $\langle softmax\rangle$ as in the original transformer, our primitive operation search space includes various kinds of unary element-wise functions and binary aggregation functions as shown in Table \ref{table: operations_attention}.
The operations such as \textit{neg, add} and \textit{multiplication}
can be performed on both scalar and matrix inputs.

\textbf{(3) Computation Graph with Variable Path Length.} As Figure \ref{fig:flow_chart} illustrates, we represent the new attention structure as a directed acyclic graph (DAG), which transforms
input nodes into the tensor output (i.e., the output of self-attentionI
layers) with multiple primitive operators in the intermediate graph.
To better promote exploration of novel attention structures, we do not fix the path length of attention computation graphs.
Note that it is possible that the dimension of the input features
in the computation graph are not matched during the calculation. We
examine whether every operation is legit and early reject those illegal
computation graphs. We also verify that the input and output dimensions
of searched attention architectures are matched to ensure layers can be stacked correctly.

\subsubsection{Inter-layer Search Space}
For the design of the whole backbone, we 1) incorporate local dependency via lightweight convolution and 2) adopt a macro search space to promote the flexibility of design.

\prj{\textbf{(1) Incorporating Local Dependencies.} As pointed out by \citep{jiang2020convbert,wu2018pay}, some of the attention heads can be replaced by local operations to better learn local dependencies as well as reduce model complexity.}
\prj{Thus, to enable a powerful and efficient language model, we consider searching a hybrid backbone to replace the attention-only architecture by adding local operations into
our inter-layer search space. Specifically, we incorporate the lightweight convolution as our candidate operation, since its effectiveness has been proven in NLP tasks such as machine
translation \citep{wu2018pay}.}

To explore whether different reception fields are preferred for different
layers, we further allow different kernel sizes
($3\times1, 5\times1, 7\times1, 9\times1, 15\times1, 31\times1, 65\times1$)
across layers. For each convolution layer, the projected input
is followed by a Gated Linear Unit (GLU) \citep{dauphin2017language},
as shown in Figure \ref{fig:flow_chart}.
\noindent

\textbf{(2) Macro Search Space.} We adopt macro search space for the backbone architecture. Specifically, we allow each layer to have different searched self-attention structure and convolution block. Comparing with the micro (cell-based) search space adopted in previous works \citep{liu2018darts, shi2020bridging}, from which a cell structure is searched and the backbone is constructed by repeatedly stacking the cell, our search space is much more flexible, which has more than $10^{20}$ possible combinations. As a result, the searched backbone architecture is more efficient and can effectively capture both global and local contexts.

\subsection{
Operation-Priority Neural Architecture Search Algorithm\label{search_algor} (OP-NAS)
}

Since we search for new architectures from scratch \prj{in an extremely large macro search space, which involves both intra-layer and
inter-layer level,} 
our NAS algorithm must be efficient, scalable, and computationally feasible. \prjj{Though gradient-based search algorithms such as DARTS
are attractive due to their search speed, they do not fit our demand 
for exploring novel attention mechanism with more flexibility. 
The supernet in gradient-based algorithms needs to store all the intermediate variables for gradient updates, 
which requires huge memory cost.  This drawback hinders their application on our search space, since we do not restrict the length of attention path and allow a large number of possible operation combinations.
} 

Evolution algorithms (EA) \citep{real2019regularized} \prj{poses less constraints over the}
search space as \prj{per} our request. However, traditional EA suffers from the risk of being trapped by local optimal in a huge search space.
To this end, we propose an operation-priority (OP) acquisition method
to improve the efficiency of model search by balancing exploration and exploitation.
Furthermore, we \prj{propose Bi-branch Weight-Sharing (BIWS) training strategy to boost model evaluation by preventing repeated pretraining.}
The details are described in Algorithm \ref{alg:operation-priority}. 

\vspace{-3mm}
\begin{algorithm}[H]
    \renewcommand\arraystretch{0.8}
    \caption{
    OP-NAS Algorithm.
    }
    \begin{algorithmic}[1]
    \STATE{Initialize population $\mathcal{M}$ from search space $\mathcal{A}$;}
    \STATE{Model evaluation in $\mathcal{M}$;}
    \REPEAT
    \STATE{\prj{$\mathcal{P}$ $\leftarrow$ Top-$K$ ($\mathcal{M}$);}}
    \FOR{each parent $p$ in $\mathcal{P}$}
    \STATE{$p'$ $\leftarrow$ $Mutation_{InterLayer}(p)$;
    }
    \STATE{$c$ $\leftarrow$ $Mutation_{IntraLayer}(p', $UCB$)$;
    }

    \STATE{Initialize $c$ with 
    \prj{BIWS} strategy
    ;}
    \STATE{\prj{Evaluate $c$ on the proxy task;}
    }
    \ENDFOR
    \STATE{\prj{
    Update $\mathcal{M}$ with the newly evaluated children.
    }
    }
    \STATE{Update UCB scores by Equation~(\ref{equ:ucb});}
    \UNTIL{convergence}
    \end{algorithmic}
    \label{alg:operation-priority}
    \end{algorithm}
\vspace{-4mm}

\subsubsection{Operation-priority Evolution Strategy}

Our OP-NAS is an evolution-based search algorithm. 
Specifically, it begins by randomly sampling candidates and evaluating them to initialize the population $\mathcal{M}$. In every iteration, the top-$K$ individuals in $\mathcal{M}$ are treated as the $parents$ to generate the $children$ via mutation. In inter-layer level, the parent follows the vanilla EA \citep{goldberg1991comparative} to perform random mutation. In intra-layer level, however, random mutation leads to severe inefficiency when searching for attention structures, as there are many possible operation combinations and the length of attention path is unconstrained.

To address the aforementioned issue, we leverage the prior information of each operation when performing intra-layer mutation.
The greedy assumption is that if a model performs well, then the operations
in its architecture (path) are promising, \prj{which should have a higher chance to be sampled. However, the algorithm should also encourage the less frequently sampled operations to prevent getting trapped in local optimal.}
Thus, we adopt the upper confidence bound (UCB) \citep{auer2002finite}
acquisition function, which balances exploitation and exploration \prj{to enhance the search efficiency and reduce the number of candidates that need to be evaluated}. 

\prjj{In contrast to previous methods which utilize acquisition functions to measure the potential of whole architectures \cite{li2017hyperband, shi2020bridging}, while the mutation is still performed randomly, our method uses the UCB acquisition function as a metric to guide the operation selection on each position during mutation. Our method is therefore more efficient and flexible, as the prior knowledge of each operation can be harnessed to generate promising children.} 
For operation $i$, the UCB score $u_{i}$ is calculated as: 
\vspace{-1mm}
\begin{small}
\begin{align}
u_{i}={\mu}_{i}+\alpha\sqrt{2\log{N}/N_{i}}\label{equ:ucb}
\end{align}
\end{small}
\hspace{-1.2mm}where \sh{${\mu}_{i}$ is the average \prj{proxy task score of the enumerated paths where operation $i$ is included, }
$\alpha$ is the hyper-parameter controlling the level of exploration, $N_{i}$ is the number of times that
operation $i$ has been sampled and $N$ is the total number of operations sampled in history.} When the operation is infrequently
sampled, the right part \prj{dominates the score function.}

\prj{As opposed to other NAS methods such as DARTS  \citep{liu2018darts} and ENAS  \citep{pham2018efficient}, whose architecture path lengths are fixed, the length of our attention path is flexible and is allowed to change during the search. Thus, assigning independent probability distributions
for operations at each position is not feasible, as the position may shift due to the change of path length. To tackle this problem, we model $n$ probability distributions, where $n$ is the length of the longest path sampled during the search. For \prj{parent} path of length $k$, the \prj{child} path is always mutated based on the first $k$ distributions.}
\prj{For convolution layers, the empirical probability distribution for different kernel sizes can be directly calculated for each layer.}
The probabilities for operations (or kernel sizes)
are calculated as: $p_{1},\dots,p_{n}=\text{softmax}(u_{1},\dots,u_{n})$,
where $u_{i}$ represents the UCB score for operation $i$.

\subsubsection{
\prj{Bi-branch Weight-Sharing (BIWS) Training Strategy}
}

To avoid the repeated pretraining of candidate models, we
design BIWS training strategy to speed up the model evaluation. Note that
even using a very reduced training scheme, evaluating one architecture by
training from scratch requires $200$ GPU hours. With our BIWS, the evaluation cost is greatly reduced by 80\%. The main idea
of our strategy is to reuse the trained model parameters
in the previous round of searching. To achieve this, we first introduce a \jh{bi-branch}
super-net which contains the largest set of the possible candidate
models: \jh{one branch contains max attention structure ($4$ input nodes), and the other branch contains the largest convolution structure
(kernel size = $65\times1$)}. Each candidate model is initialized by the parameters fetched from the corresponding layers and positions of the super-net. \prj{In this way, we can obtain evaluation results with high fidelity after only a few epochs of fine-tuning.}
To enable a reusable super-net, we design the following strategies: 

\begin{figure}[h]
\centering
 	\vspace{-3mm}
	\includegraphics[width=0.48\textwidth]{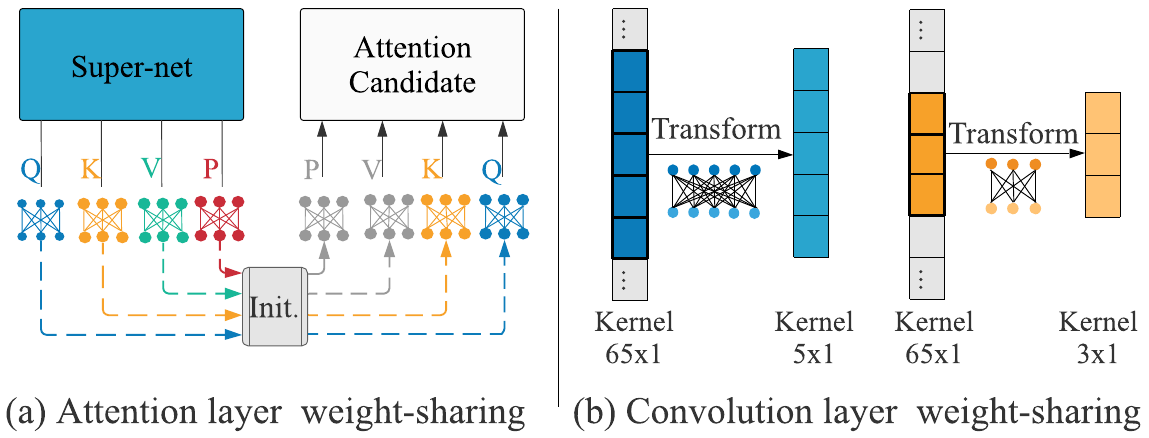}
    \vspace{-5mm}
	\caption{Illustration of BIWS strategy. For attention layers, transformation matrices of K,Q are initialized from corresponding positions of the largest $4$-node attention. For convolution layers, small kernels are initialized by the center of the largest kernel.}
	\label{fig:weight-sharing}
	\vspace{-4mm}
\end{figure}

\textbf{(1) Convolution layer weight-sharing.} Inspired by \citep{cai2019once},  we maintain the weights for the largest convolution layer (kernel size $=65\times1$) throughout searching, then the weights at the center position are shared to initialize the small kernels for the candidate models (as shown in Figure~\ref{fig:weight-sharing}). Since the shared weights play multiple roles when they are applied to sub-kernels of various sizes, the weights in those sub-kernels should have different properties of distributions and magnitudes. To this end, we introduce the kernel transformation matrices to adapt the shared weights for sub-kernels of different sizes. 
{Specifically, different kernel transformation matrices are learnt during training for different layers, while being shared across all the channels within each layer.} 
The weights of the sub-kernels
are updated to the largest kernel in the super-net after training the candidate models in each round.

\textbf{(2) Attention layer weight-sharing.} The parameters in self-attention
structure lie in the linear transformation matrices for \textit{key,
query, value} and $P$. Since we only mutate parts of the computation
graph in each round of searching, we can directly initialize these fully-connected
layers in the child individuals using the weights extracted from the corresponding
layers of the super-net. 

\begin{table*}[t]
\vspace{-4mm}
\begin{center}
\centering{}
\resizebox{\textwidth}{!}{
\begin{tabular}{lccccccccccc}
\hline
& \#Params & Infer FLOPs & CoLA & MRPC &  MNLI-(m/mm) &  STS-B & RTE &QQP &QNLI &SST-2 & AVG\tabularnewline
\hline
\multicolumn{12}{l}{\textbf{\textit{Development Set}}} \tabularnewline
BERT-base(ours) &110M & 2.9e10 & 58.1  & 89.7   & 84.8/85.2  & 88.8 &69.0 &88.2 & 91.5 &92.9&83.1 \tabularnewline
AutoBERT-att& 104M& 2.3e10 &   65.4  & 92.2  & 84.6/85.0  &90.4 &81.6& 88.5 & 91.8& 93.8 & 85.9  \tabularnewline
AutoBERT-conv& 104M& 2.2e10 & 63.8 & {92.6}  & 84.4/84.6  &90.1 & 80.5  & 88.3&91.7& 93.5 &85.5 \tabularnewline
AutoBERT-w/o-desc& 104M  & 2.3e10  &65.1  &  {92.8}  & 84.5/85.0  & 90.5 &78.7 & 88.2 &91.6 &93.7 & 85.6\tabularnewline
AutoBERT-Zero& 104M & 2.3e10 &  {64.5}  &  {93.3}  &  {85.5/85.3}  &  {90.8}& {81.9} &{88.9} & 92.0 & {94.2}&  {86.3} \tabularnewline
AutoBERT-Zero$^\ast$ & 104M &2.3e10 & \textbf{67.3} & \textbf{93.8} & \textbf{86.4/86.3} & \textbf{90.8} & \textbf{85.2} & \textbf{91.7} & \textbf{92.5} & \textbf{95.2} & \textbf{87.7}\tabularnewline

\hline
\hline
\multicolumn{12}{l}{\textbf{\textit{Test Set}}}
\tabularnewline
GPT\citep{radford2018improving} & 117M & 3.0e10 & 45.4 & 82.3 & 82.1/81.4 & 82.0 & 56.0 & 70.3 & 88.1 & 91.3 & 75.4 \tabularnewline
BERT-base\cite{devlin2019bert} & 110M  & 2.9e10 & 52.1  &88.9   & 84.6/83.4  & 85.8 &66.4 &71.2 & 90.5 &93.5&79.6 \tabularnewline
DynaBERT-base\cite{DBLP:conf/NeurIPS/HouHSJCL20} & 110M  & 2.9e10 & 54.9 & 87.9 &84.5/84.1 &84.4 &69.9 &{72.1} &{91.3} &93.0 & 80.2 \tabularnewline
ConvBERT-base \citep{jiang2020convbert} & 106M & 2.7e10  & 53.7 & 89.3 & 84.6/83.6 & 86.1 & 72.1 & 71.3 & 90.1 & 93.5 & 80.5 \tabularnewline
Roberta-base \citep{liu2019roberta} & 110M & 2.9e10 & 50.5 & 90.0 & 86.0/85.4 & 88.1 & 73.0 & 70.9 & 92.5 & 94.6 & 81.1 \tabularnewline
BERT-Large\cite{devlin2019bert} & 340M & 8.7e10 & 60.5 & 89.3 & 86.7/89.5 & 86.5 & 70.1 & 72.1 & 92.7 & 94.9 & 82.1 \tabularnewline
\hline
AutoBERT-Zero & 104M & 2.3e10 & {55.9} & {89.5} &{85.4/84.9} &{88.3} &{77.8} &71.8 &91.2& {94.6}  & {82.2} \tabularnewline
AutoBERT-Zero$^\ast$ & 104M & 2.3e10 & \textbf{59.5} & \textbf{90.5} & \textbf{86.1/86.0} & \textbf{88.9} & \textbf{80.2} & \textbf{72.8} & \textbf{92.1} & \textbf{95.1} & \textbf{83.5} \tabularnewline
AutoBERT-Zero-Large & 318M & 6.8e10 & \textbf{63.8} & \textbf{90.7} & \textbf{87.7/87.1}  & \textbf{90.1} & \textbf{80.4} & \textbf{72.1} & \textbf{93.6} & \textbf{95.4} & \textbf{84.5} \tabularnewline
\hline
\end{tabular}
}
\caption{Performance comparison on the test set of GLUE.
Our 12-layer base model AutoBERT-Zero significantly surpasses RoBERTa-Base and BERT-large (24 layers). Note that Roberta \citep{liu2019roberta} runs on 160G corpus, whereas our model runs on 16G corpus. \jh{Infer FLOPs assumes single inputs with length 128. 
AutoBERT-Zero$^\ast$ is initialized from the surpernet.}}\label{tab:compare_on_glue_test} 
\end{center}
\end{table*}

\vspace{-1mm}
\section{Experiments}\label{sec:experiments}
\subsection{Dataset and Setting}\label{sec:implementation}
\textbf{Datasets and metrics.} We first pre-train the backbone architectures using a large corpus of text data and then finetune the model for each specific downstream task. For pre-training, we use the BooksCorpus \citep{zhu2015aligning} and English Wikipedia \citep{devlin2019bert}.
For finetuning and evaluation, we use the General Language Understanding Evaluation (GLUE) \citep{wang2018glue}
and the Stanford Question Answering Dataset (SQuAD) \citep{rajpurkar-etal-2016-squad}.
Unless stated otherwise, downstream tasks  are reported using the same metrics in BERT \citep{devlin2019bert}.
For other settings, we follow the settings of BERT paper.

\textbf{Implementation Details.} 
We use Masked Language Model (MLM) and Next Sentence Prediction (NSP) as pre-training tasks.
The whole process can be divided into two phases, namely the NAS phase and the fully-train phase. For NAS phase, we train the base model, whose configuration is the same as BERT-base ($L=12,H=768,A=12$).
Initial {$\mathcal{M}$} is set as 100, and $K$ is set as 5.
Each parent will mutate 5 child architectures. In the NAS phase, we train each candidate architecture for 40,000 steps, which is then evaluated on the proxy task (GLUE).
The searching phase costs around 24K GPU hours (760+ evaluated candidates) on Nvidia V100. If we only use EA without BIWS strategy, the computation cost is estimated to be about 182K GPU hours. 
In fully-train phase, we first pre-train the searched base-size model. To further verify the model’s scaling ability, we also fully-train the model on small model ($L=12,H=256,A=4$) and large model ($L=24,H=1024,A=16$).
\jh{Specifically, we treat each two continuous layers as a block and expand the base model to large model by inserting the same block after the original one.}
More details are attached to Appendix. 

\subsection{Results and Analysis}\label{sec:analysis}
\begin{figure}[t]
\centering
 	\vspace{-5mm}
	\includegraphics[width=0.48\textwidth]{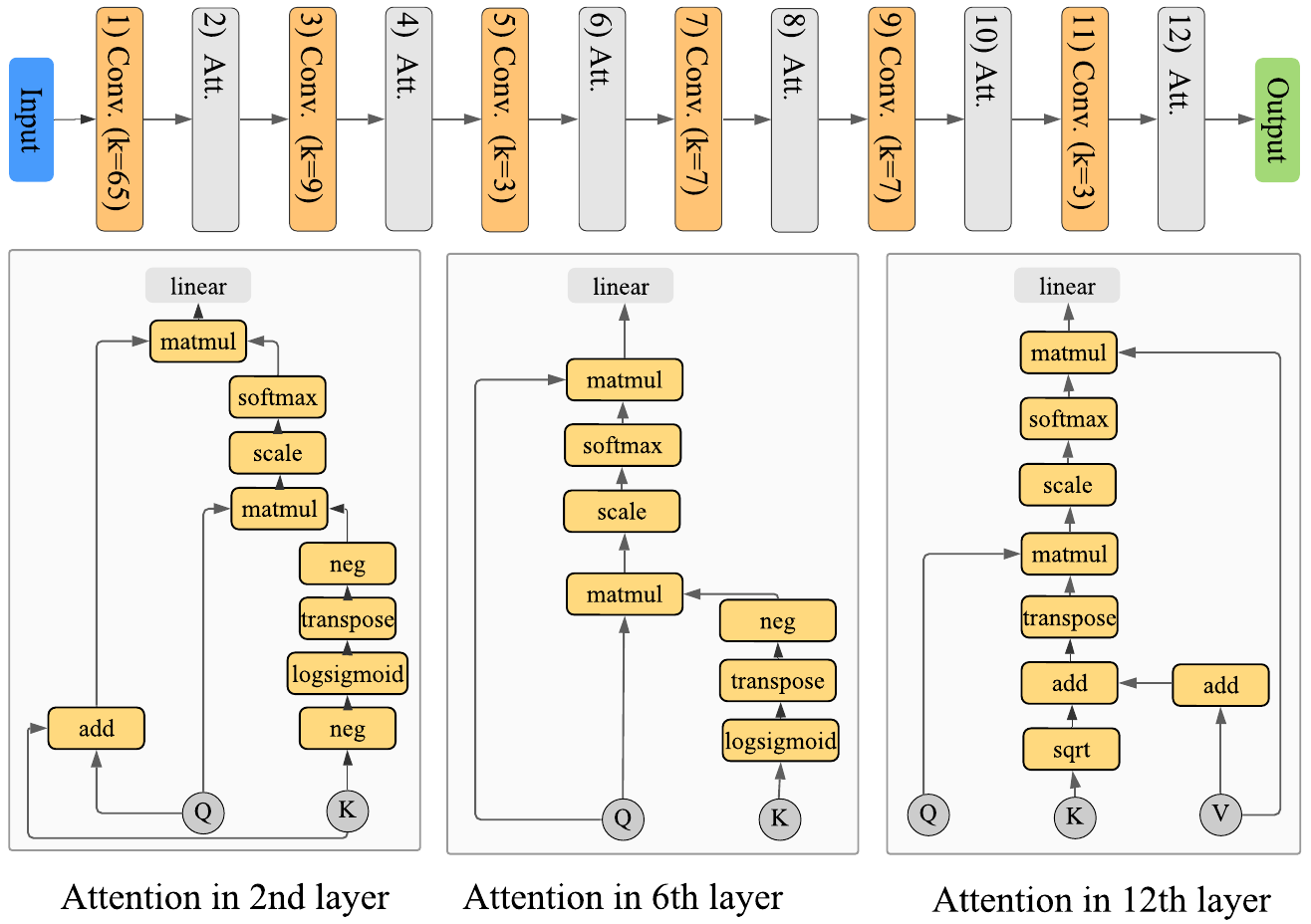}
	\vspace{-6mm}
	\caption{The detailed architecture of AutoBERT-Zero. We only show the 2\textsuperscript{nd}, 6\textsuperscript{th} and 12\textsuperscript{th} discovered attention structures due to limited space. Att. and Conv. represents the searched attention layer and convolution layer respectively. The whole backbone is attached to the Appendix.
	}
	\label{fig:autobert-zero}
 	\vspace{-5mm}
\end{figure}

\textbf{Structure Analysis of AutoBERT-Zero.} We name the best searched architecture of OP-NAS \textbf{AutoBERT-Zero}.
As shown in Figure~\ref{fig:autobert-zero},  
the hybrid backbone of AutoBERT-Zero is constructed with stacked conv-att blocks (searched convolution followed by searched attention layer), which effectively integrates the local and global dependencies of natural language.
For the searched attentions, 
$V$ is shared with $Q$/$K$ in shallow layers, but non-shared in the deeper layers. This is reasonable since the shallow layer only process the low-level features, whereas the deep layers need more parameters to capture the complex semantic features. For example, 
$\hat{Attn}(X)_{L_2}$ introduces $K$-$V$ and $Q$-$V$ sharing mechanisms, while $\hat{Attn}(X)_{L_{12}}$ adopts separate weights for $K$, $Q$ and $V$:
\vspace{-3mm}
\begin{small}
\begin{align*}
    \hat{Attn}(X)_{L_2}
    &=\sigma(Q\log(1+\exp(K^\top))/\sqrt{d_h})(K+Q)W_{O}^{\top}.\\
    \hat{Attn}(X)_{L_{12}}
    &=\sigma(Q(K/\sqrt{d_h}+V)^\top/\sqrt{d_h})VW_{O}^{\top}. 
\end{align*}
\end{small}
\hspace{-2mm}
Besides, the kernel sizes of convolution layers roughly follow a descending order (changing from 65 to 3), which indicates the convolution layers learn local information from wide to narrow. \prj{This is justifiable as the a  larger receptive field captures more information, which helps emphasize on the informative features while suppress the unimportant ones. After the shallower layers effectively reduce the information redundancy, the deeper layers can focus on the important semantic features.}

\begin{table}[t]
\begin{minipage}[c]{0.48\textwidth}
\vspace{-5mm}
\begin{center}
\vspace{-3mm}
\centering{}
\renewcommand\arraystretch{1.15}
\resizebox{1.0\textwidth}{!}{
\begin{tabular}{lccccccccc}
\hline
& \multirow{2}{*}{\#Params of Att}& \multicolumn{2}{c}{SQuAD v1.1} & \multicolumn{2}{c}{SQuAD v2.0} \tabularnewline
& & EM & F1 & EM & F1    \tabularnewline
\hline
BERT-base(ours) & 21.3M & 78.9  & 86.7  &  70.2 & 72.8  \tabularnewline
AutoBERT-att & 15.9M &  79.7 &  87.5 & \textbf{72.9} & \textbf{75.7}  \tabularnewline
AutoBERT-conv& 15.4M & 79.1  & 86.5  & 71.9  &  74.6   \tabularnewline
AutoBERT-w/o-desc& 15.4M& 79.5 & 87.0 & 71.5 & 73.9    \tabularnewline
AutoBERT-Zero & 15.4M & \textbf{79.9}  & \textbf{87.6}  & {72.5} &{75.0} \tabularnewline
\hline
\end{tabular}}
\vspace{-2mm}
\caption{\small{Results on SQuAD(dev). ``\#Params of Att" counts parameters in attention structures.
} 
\label{tab:Comparison-of-squad}}
\end{center}
\vspace{+1mm}
\end{minipage}
\hspace{2mm}
\begin{minipage}[c]{0.48\textwidth}
\begin{center}
\centering{}
\vspace{-2mm}
\resizebox{1.0\textwidth}{!}{
\begin{tabular}{lcccccccccc}
\hline
& \#Params & FLOPs & Pre-train Task  & GLUE \tabularnewline
\hline
ELMO \cite{Peters:2018} & 96M & 2.6e10 & LM & 71.2 \tabularnewline
GPT\cite{radford2018improving} & 117M & 3.0e10 & LM & 78.8 \tabularnewline
BERT-small \citep{jiang2020convbert} & 14M & 3.7e9 & MLM & 75.1 \tabularnewline
ELECTRA-small \citep{clark2019electra} & 14M & 3.7e9 & RTD & 79.9 \tabularnewline
ConvBERT-small \citep{jiang2020convbert} & 14M & 4.1e9 & MLM & 75.9 \tabularnewline
AutoBERT-Zero-small & 13M & 2.9e9 & MLM & \textbf{80.5} \tabularnewline
\hline
BERT-large\citep{devlin2019bert} & 340M & 8.7e10 &MLM & 84.4 \tabularnewline
AutoBERT-Zero-large & 318M & 6.8e10 & MLM & \textbf{87.9} \tabularnewline
\hline
\end{tabular}}
\vspace{-2mm}
\caption{\small{Scaling ability of the searched model. Results are reported on GLUE dev set.\protect\footnotemark[2]} \label{tab:small_models}}
\end{center}
\vspace{-7mm}
\end{minipage}
\end{table}
\footnotetext[2]{Following ConvBERT,  we count accuracy for MRPC and QQP
for small model. 
Small model results are median results of 3 runs.}

\begin{table*}[t]
\begin{center}
\centering{}
\renewcommand\arraystretch{0.93}{
\begin{tabular}{lcccccccccc}
\hline
& \#Params of Att & CoLA & MRPC &  MNLI-(m/mm) &  STS-B & RTE &QQP &QNLI &SST-2 & AVG\tabularnewline
\hline
BERT-base &21.3M & 58.1  & 89.7   & 84.8/85.2  & 88.8 &69.0 &88.2 & 91.5 &92.9&83.1 \tabularnewline
Att-only & 16.5M & 60.0  &  92.1 &  84.9 /84.1 & 90.6  & 79.4 & 88.3 & 91.5 & 92.5 & 84.8 \tabularnewline
Conv-only  & 15.4M & 53.7  & 82.9  & 69.0/66.1  & 81.0  & 64.2 &82.0 & 75.7 &86.7 &73.3 \tabularnewline
AutoBERT-Zero & 15.4M & \textbf{64.5}  & \textbf{93.3}  & \textbf{85.5/85.3}  &\textbf{90.8}   &\textbf{81.9} & \textbf{88.9} & \textbf{92.0} & \textbf{94.2} & \textbf{86.3} \tabularnewline 
\hline
\end{tabular}}
\vspace{-2mm}
\caption{Model comparison among AutoBERT-Zero and its Attention-only, Conv-only variants. Models are fully-trained and evaluated on GLUE dev set. \label{tab:Comparison-of-different}}
\end{center}
\vspace{-4mm}
\end{table*}

\textbf{Results on GLUE \& SQuAD.} After the NAS phase, the searched models are fully-trained and evaluated on downstream tasks. \prj{Our AutoBERT-Zero consistently outperforms other baselines by a large margin. To demonstrate the superiority of AutoBERT-Zero's structure, we fully-train several other searched backbones for comparison:}
\Ni \textbf{AutoBERT-w/o-desc.} A backbone without descending kernel sizes for convolution layers.
\Nii \textbf{AutoBERT-att.} A backbone containing three continuous attention layers.
\Niii \textbf{AutoBERT-conv.} A backbone containing three continuous convolution layers. 
The details of architectures can be found in Appendix. As shown in Table~\ref{tab:compare_on_glue_test}, AutoBERT-Zero achieves the highest GLUE score, with a significant performance gain over BERT-base while having less parameters and FLOPs. Specifically, AutoBERT-Zero performs much better than AutoBERT-att and AutoBERT-conv, demonstrating that the conv-att block can better integrate the local and global dependencies. Besides, AutoBERT-Zero's advantage over AutoBERT-w/o-desc indicates that the kernel size pattern from wide to narrow in convolution layers benefits the performance. As shown in Table~\ref{tab:Comparison-of-squad}, AutoBERT-Zero consistently surpasses BERT-base on both SQuAD v1.1 and v2.0, demonstrating the generalizibility of our searched model.

\begin{figure}[t]
\centering
	\vspace{-4mm}
	\includegraphics[width=0.38\textwidth]{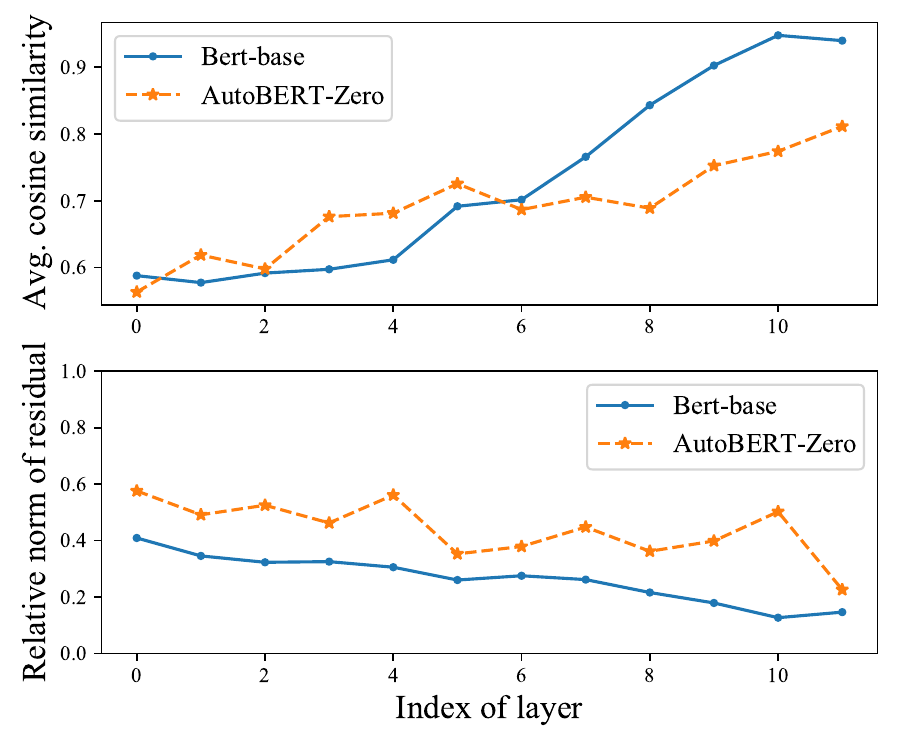}
	\vspace{-3mm}	
	\caption{\small{Oversmooth exists in purely-stacked transformer structure, whereas the hybrid structure helps to tackle this problem.}}
	\vspace{-5.5mm}
	\label{fig:cosine_rank}
\end{figure}

\textbf{Representation ability of AutoBERT-Zero.} ``Token-uniformity'' damages model's representation ability. To measure the degree of ``token-uniformity'', following \citep{dong2021attentionnot,gong2021improve}, 
we use \textit{relative norm of residual
} to measure the rank of output ( the rank is equal to 1 when \textit{residual} is equal to 0), and measure the average \textit{pairwise cosine-similarity} between the representations of different tokens on 1,280 samples of STS-B. As shown in Figure~\ref{fig:cosine_rank}, latent representations from purely-stacked BERT-base have high similarity, and the rank of output is closer to 1 (\textit{relative norm of residual} is closer to 0), showing no significant difference between the tokens. On the other hand, the output of AutoBERT-Zero has relatively larger \textit{residual} and lower token similarity, showing that the hybrid backbone helps mitigate this problem.

\textbf{Scaling ability of AutoBERT-Zero.}
We further extend AutoBERT-Zero structure to different capacities, 
showing strength in both large and small models. Specifically, Table~\ref{tab:small_models} shows that our large model surpasses BERT-large by 3.5 in GLUE. 
Remarkably, our small model significantly surpasses the SOTA ConvBERT-small (4.6 higher) and BERT-small (5.4 higher) under the vanilla MLM task. Besides, our small model considerably outperforms the large GPT in terms of both performance and complexity:  1.7 higher GLUE, 88\% less parameters, and 90\% less FLOPs. 
Despite the advantage of strong pre-train task (RTD), ELECTRA-small is still outperformed by our model.

\begin{figure}[t]
\vspace{-4mm}
\centering
\includegraphics[width=0.38\textwidth]{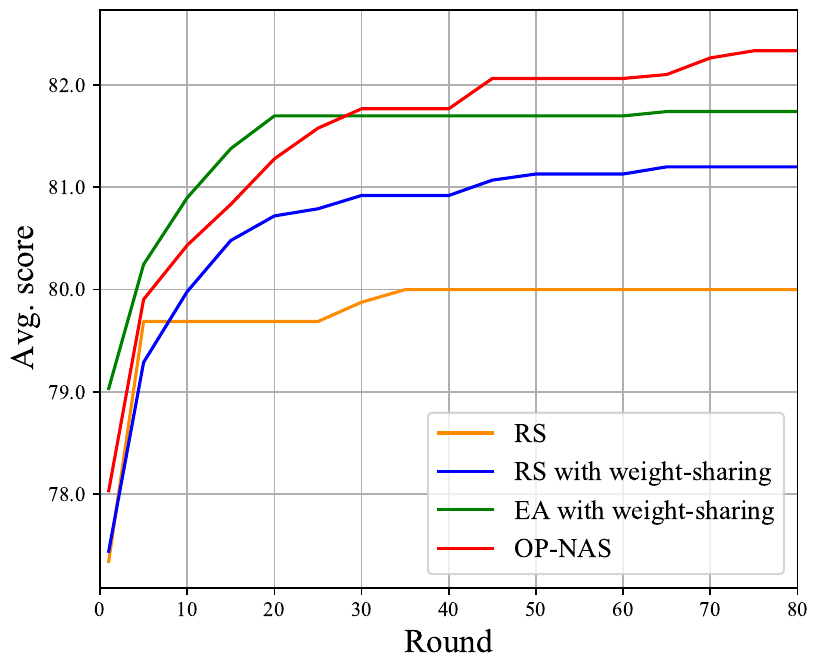}
\vspace{-2mm}	
\caption{\small{Searching performance comparison among Random Search (RS), RS with weight sharing, EA with weight sharing and OP-NAS.}}
\label{fig:search_algorithms}
\vspace{-5.5mm}
\end{figure}

\textbf{The Efficiency of OP-NAS.}  During the search, we observe that by adopting the proposed operation-priority strategy, the exploration ability of the EA is highly improved, which prevents getting trapped in local optimal (see Figure~\ref{fig:search_algorithms}). The results shows that searched model using OP-NAS outperforms other NAS algorithms by a large margin.
\prj{As the quality of model evaluation during NAS phase greatly impacts the algorithm's effectiveness, we further examine the fidelity of the evaluation results. Kendall~\citep{kendall} correlation analysis is performed to evaluate the correlation between model performances in the NAS phase and fully-train phase.} As shown in Appendix B, \prj{high}
correlations are captured in most of the downstream tasks,
\prj{which is owing to the effectiveness of our BIWS strategy}.

\textbf{Ablation study.} To investigate the superiority of searched hybrid architecture, \prj{
we evaluate performance of attention-only and convolution-only variants, which are constructed by stacking either the searched attention or the convolution layers of AutoBERT-Zero.} 
\prj{For example, for the attention-only variant, each convolution block is replaced with the attention layer directly behind it.} From Table~\ref{tab:Comparison-of-different}, we find that the hybrid backbone architecture outperforms both attention-only and convolution-only variants. Besides, the attention-only variant surpasses BERT-base by a large margin, showing effectiveness of searched attention structures.

\vspace{-2mm}
\section{Conclusion}\label{sec:conclusion}
\prj{In this work, we propose a novel hierarchical search space and an efficient NAS framework to automatically find promising PLM backbones from scratch, which prevents the tedious manual tuning.  The searched self-attention structure and backbone architecture can inspire new insights for model design in the NLP community. 
}

\section*{Acknowledgments}
We would like to thank Renjie Pi and the anonymous reviewers for insightful suggestions that have significantly improved the paper. 
The research of Philip L.H. Yu was supported by a start-up research grant from the Education University of Hong Kong (\#R4162).

\section*{Appendix}

\appendix

\begin{figure*}[t]
\centering
	\includegraphics[width=0.9\textwidth]{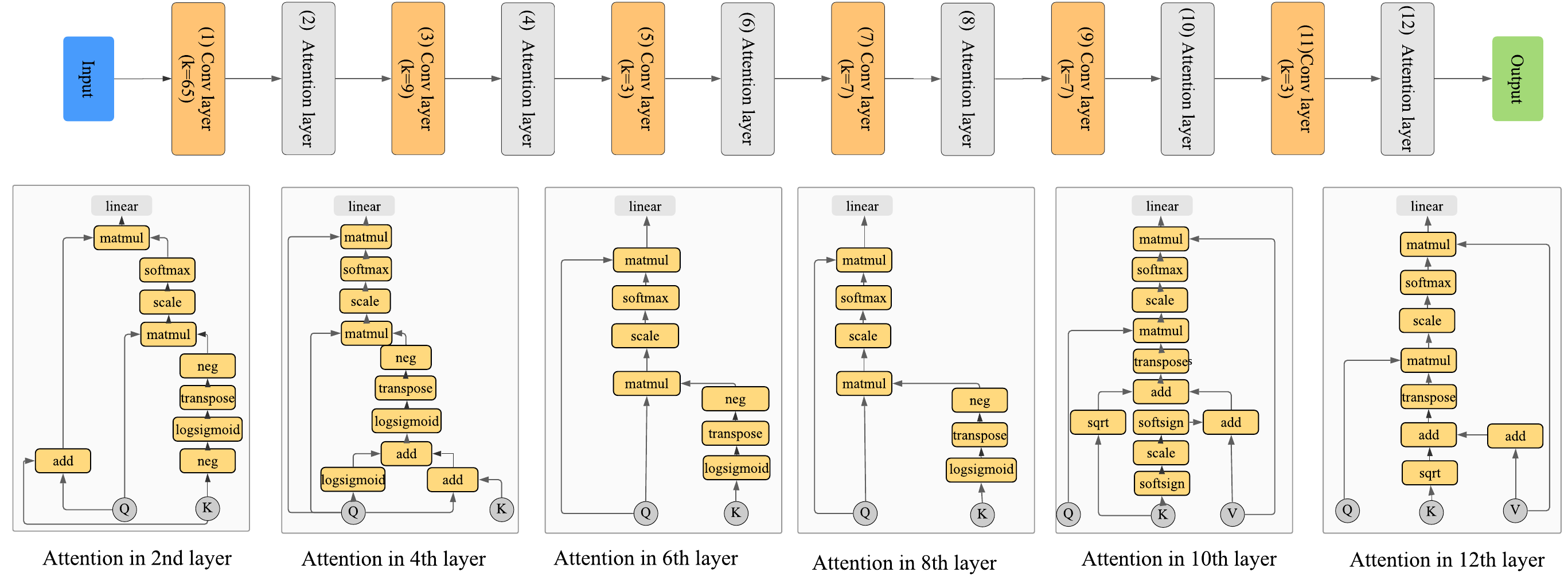}
	\vspace{-3mm}
	\caption{The detailed architecture of \textbf{AutoBERT-Zero}. This hybrid backbone is constructed with stacked conv-att blocks (searched convolution followed by searched attention layer), which effectively integrates the local and global dependencies of natural language. For the searched attentions, $V$ is shared with $Q$/$K$ in shallow layers, but non-shared in the deeper layers. }
	\label{fig:autobert-zero}
\end{figure*}

\begin{figure*}[t]
\centering
	\vspace{-2mm}
	\includegraphics[width=0.8\textwidth]{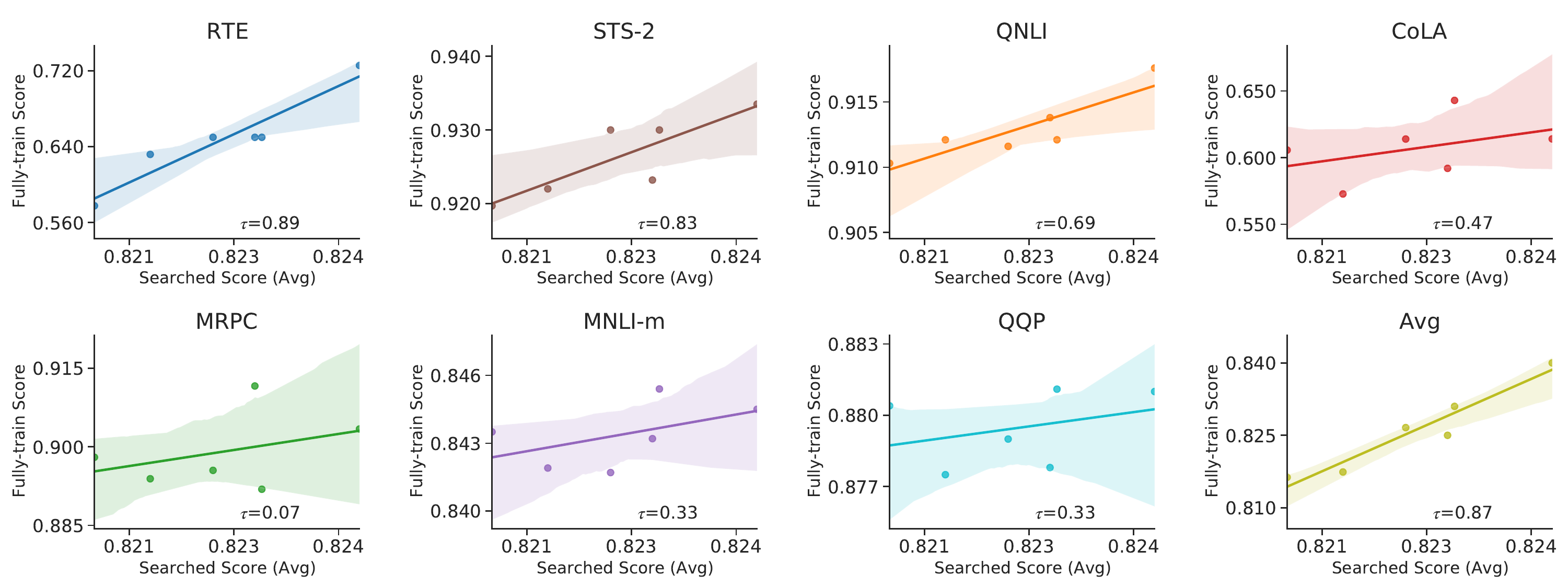}
	\vspace{-3mm}	
	\caption{
	High
	correlation can be found between the searched results and final results on different GLUE tasks. Six architectures of the candidates generated by OP-NAS are randomly selected for comparison. }
	\vspace{-3mm}
	\label{fig:corr_analysis}
\end{figure*}

\section{Detailed searched architectures}
The detailed architetcture of best searched model AutoBERT-Zero is shown in Figure~\ref{fig:autobert-zero}. The hybrid backbone of AutoBERT-Zero is constructed with stacked conv-att blocks (searched convolution followed by searched attention layer), which effectively integrates the local and global dependencies of natural language. 

\begin{figure*}[t]
\begin{minipage}[c]{\textwidth}
\centering
	\includegraphics[width=0.9\textwidth]{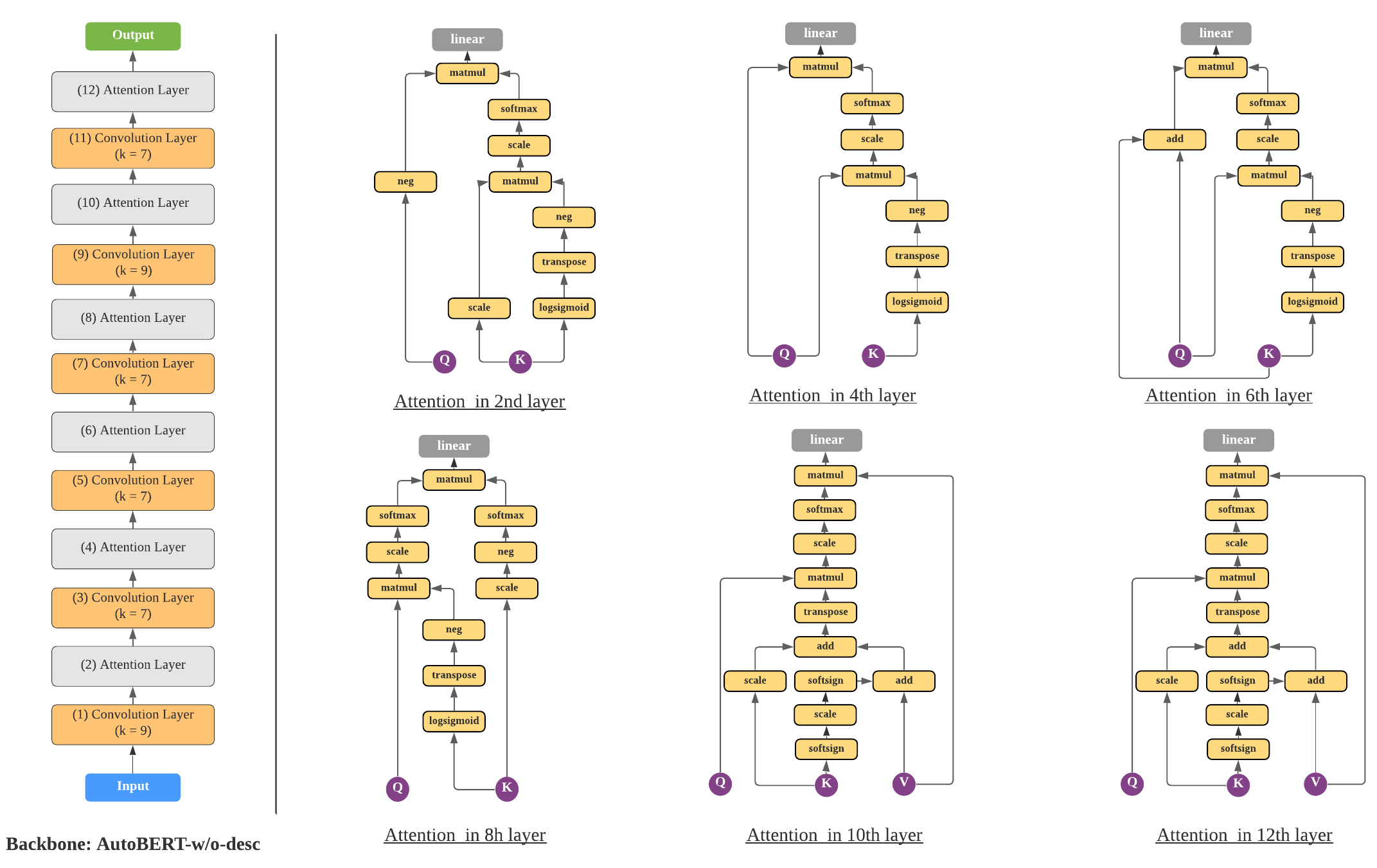}
	\caption{The detailed architecture of \textbf{AutoBERT-w/o-desc}. This model is similar to AutoBERT-Zero, but its convolution layers' kernel sizes are not following a descending order. The architectures of attention layers also become more complex as the layers go deeper.}
	\label{fig:autobert-zero-wo-desc}
\end{minipage}
\begin{minipage}[c]{\textwidth}
\centering
	\includegraphics[width=0.9\textwidth]{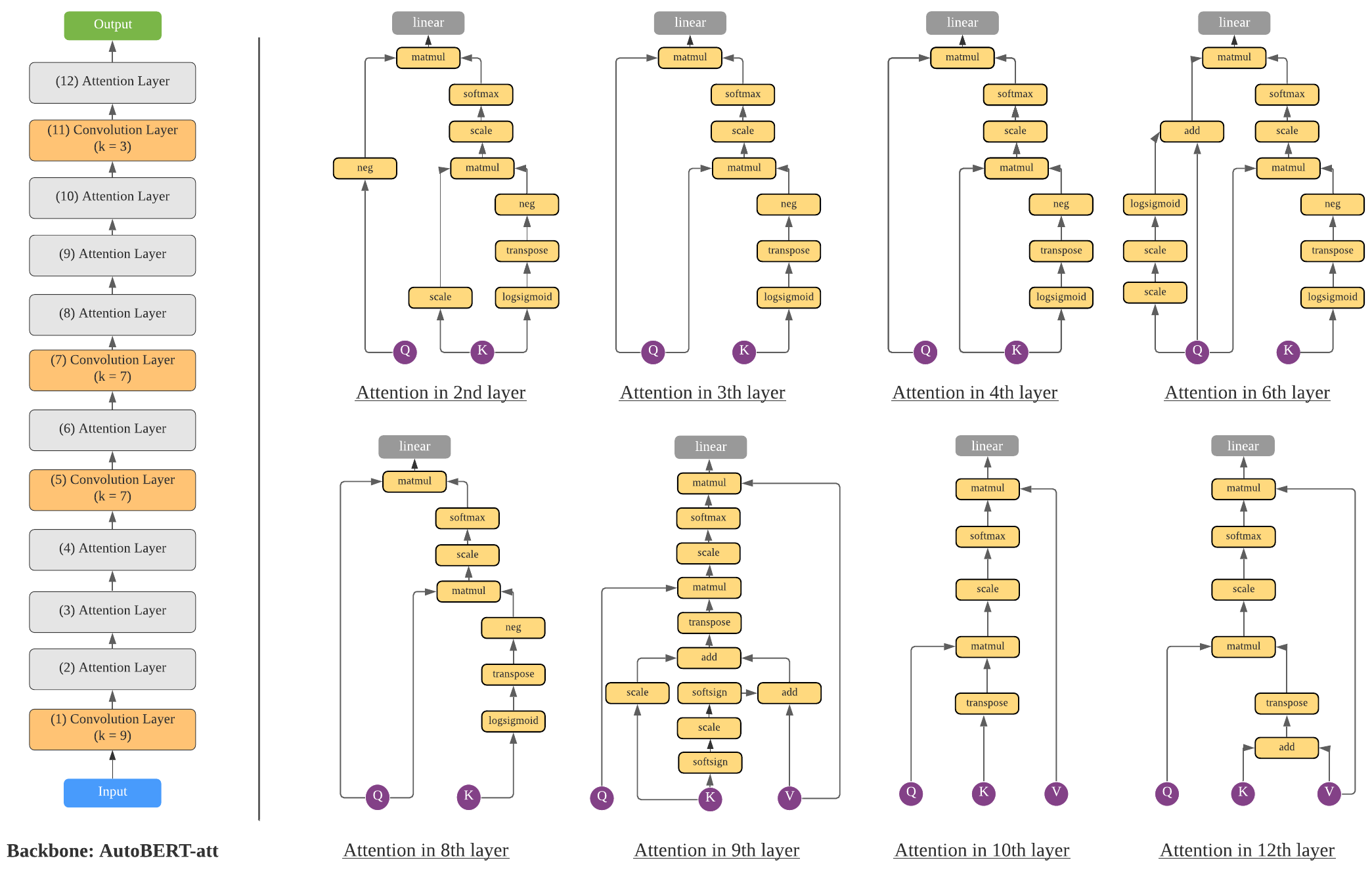}
	\caption{The detailed architectures of \textbf{AutoBERT-att}. This model has several continuous attention layers. Like other backbones, the deeper layers of this backbone also have more input nodes.}
	\label{fig:autobert-att}
\end{minipage}
\end{figure*}

\begin{figure*}[h]
\centering
	\includegraphics[width=0.9\textwidth]{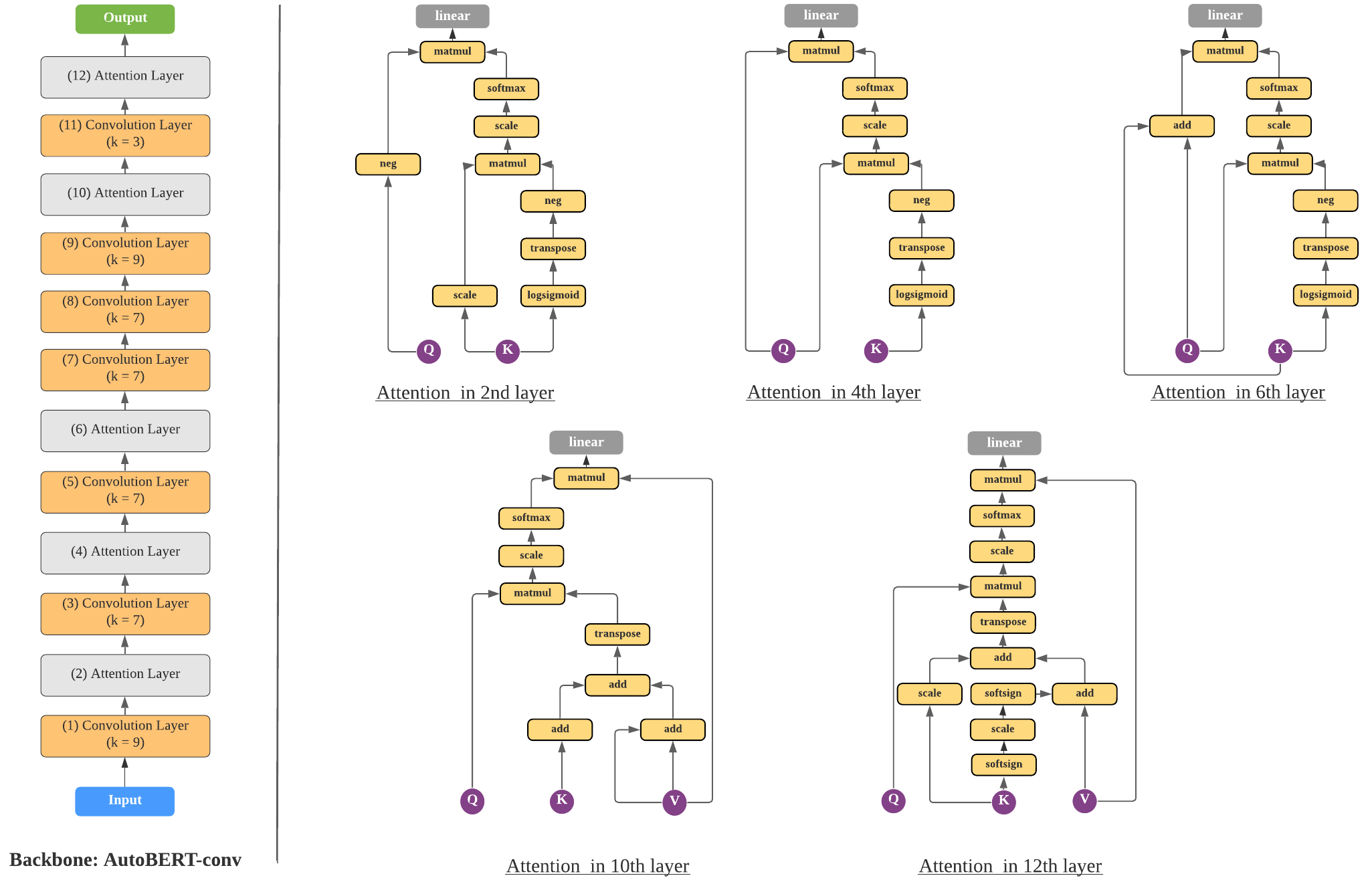}
	\caption{The detailed architecture of \textbf{AutoBERT-conv}. This model contains several continuous convolution layers, whose kernel sizes do not follow an obvious order (ascending/descending). However, the deeper layers also have more input nodes in the attention modules.}
	\label{fig:autobert-conv}
\end{figure*}

Apart from the best searched model AutoBERT-Zero (see Figure 5 in the main paper), we randomly choose several searched architectures from the final searching round as examples. Detailed architectures are shown on Figure~\ref{fig:autobert-zero-wo-desc} $\sim$ \ref{fig:autobert-conv}. Due to space limit, these figures are shown in the last pages.

\section{Correlation between searching and fully-train phases}
As the quality of model evaluation during NAS phase greatly impacts the algorithm’s effectiveness, we further examine the fidelity of the evaluation results. Figure~\ref{fig:corr_analysis} shows the kendall~\citep{kendall} analysis between model performances in NAS phase and fully-train phase.   We can find high correlation between results in the searching phase and in the fully-train phase on various downstream tasks, showing model's performance in searching phase can effectively represent the its 
quality in fully-train phase, which further verifies the BIWS strategy's effectiveness in pre-training a model in a short time (40,000 updates).

\section{Datasets}
GLUE is a collection of diverse natural language understanding datasets. Unless stated otherwise, follow \citep{devlin2019bert}, we use the accuracy to evaluate MNLI, QNLI, RTE, SST, use F1 to evaluate MRPC and QQP, use Spearman correlation for STS-B and Matthews correlation for CoLA. The average score is denoted by GLUE. 

SQuAD is a question answering dataset containing 100k question/answer pairs. The target of this task is to locate the answer with the given context and questions. The Exact Match and F1 scores are reported for SQuAD. We provide further details about GLUE and SQuAD tasks below.

\subsection{GLUE}
\paragraph{MNLI.}
A ternary classification task: Given a premise sentence and a hypothesis sentence, the target of the Multi-Genre Natural Language Inference (MNLI) \citep{williams2018broad} is to predict whether the last sentence is an entailment, contradiction, or neutral relationships with respect to the first one.

\paragraph{QQP.}
A binary classification task:  Given two questions from Quora, the target of Quora Question Pairs (QQP) \citep{chen2018quora} is to determine whether these two asked questions are semantically equivalent or not.

\paragraph{QNLI.}
A binary classification task: The Question Natural Language Inference (QNLI) \citep{wang2018multi} is derived from the Stanford Question Answering Dataset \citep{rajpurkar2016squad}. Given a sentence pairs (question, sentence), the target of QNLI is to predict whether the last sentence contains the correct answer of the question.

\paragraph{SST-2.}
A binary classification task: The Stanford Sentiment Treebank (SST-2) \citep{socher2013recursive} aims to predict the sentiment for a single-sentence. All sentences are extracted from movie reviews with human annotations of their sentiment.

\paragraph{CoLA.}
A binary classification task: The Corpus of Linguistic Acceptability (CoLA) \citep{warstadt2019neural} is consisting of English acceptability judgments extracted from books and journal articles. Given a single-sentence, the target is to determine whether the sentence is linguistically ``acceptable" or not.

\paragraph{STS-B.}
A quinary classification task: The Semantic Textual Similarity Benchmark (STS-B) \citep{cer2017semeval} aims to predict the similarity score (from $1$ to $5$) between a given sentence pair, whose sentence pairs are drawn from news headlines and other sources.

\paragraph{MRPC.}
A binary classification task: The Microsoft Research Paraphrase Corpus (MRPC) \citep{dolan2005automatically} consists of $5,801$ sentence pairs automatically extracted from online news sources, with human annotations for whether the sentences in the pair are semantically equivalent.

\paragraph{RTE.}
A binary classification task: Recognizing Textual Entailment (RTE) is similar to MNLI aiming to predict the entailment, but with much less training data \citep{dagan2005pascal}.

\subsection{SQuAD}
\paragraph{SQuAD v1.1.}
The Stanford Question Answering Dataset (SQuAD v1.1) \cite{rajpurkar2016squad} is a large-scale question and answer task consisting of $100k$ question and answer pairs from more than $500$ articles. Given a passage and the question from Wikipedia, the goal is to determine the start and the end token of the answer text.

\paragraph{SQuAD v2.0.}
The SQuAD v2.0 task is the extension of above SQuAD v1.1, which contains the $100k$ questions in SQuAD v1.1 and $50k$ unanswerable questions. The existence of unanswerable question makes this task more realistic and challenging.

\begin{table}[h]
\vspace{-2mm}
\vspace{-3mm}
\begin{center}
\resizebox{0.48\textwidth}{!}{
\begin{tabular}{lccc}
\hline & Small & Base & Large  \tabularnewline
\hline
Layer & 12  & 12 & 24  \tabularnewline 
Hidden dim & 256 & 768 & 1024 \tabularnewline
Word embedding dim & 256 & 768 & 1024\tabularnewline
Intermediate layer dim & 1024 & 3072 &4096 \tabularnewline
Attention heads & 4 & 12 & 16 \tabularnewline
Head dim & 64 & 64& 64 \tabularnewline
Learning rate & 1e-4 & 1e-4 & 1e-4 \tabularnewline
Learning rate decay & Linear &Linear & Linear \tabularnewline
Warm-up proportion & 0.01 & 0.01 & 0.02\tabularnewline
Adam $\beta_{1}$ & 0.9 & 0.9 & 0.9  \tabularnewline
Adam $\beta_{2}$ & 0.999 & 0.999 & 0.999 \tabularnewline
Dropout & 0.1 & 0.1 & 0.1 \tabularnewline
Batch size & 320 & 160 & 80  \tabularnewline
Input sequence length  & 128 & 128 &128\tabularnewline
\hline
\end{tabular}}
\caption{Hyperparameters for pre-training.}\label{tab:pretrain_implementation_details}
\end{center}
\vspace{-5mm}
\end{table}

\section{Implementation details}

The whole process can be divided into two phases, namely the NAS phase and the fully-train phase. For both phases, we pre-train our base model, whose configuration is the same as BERT-base($L=12,H=768,A=12$). More specifically, we set the number of transformer layers $L=12$, the hidden state size $H=768$, the number of heads in each layer $A=12$ and the number of neurons in the intermediate layer $d_{ff}=3072$ (4 times of the hidden dimension).

\subsection{Details in Searching Phase}
\paragraph{Model Searching.}
For the model searching phase, we proposed the Operation-Priority(OP) strategy to generate new model candidates. Initial {$\mathcal{M}$} is set as 100, and $K$ is set as 5. The confidence value $c$ in the UCB function is set as 0.01. Each parent will mutate 5 child architectures. The searching phase costs around 24K GPU hours (760+ evaluated candidates) on Nvidia V100. If we only use EA without BIWS strategy, the computation cost is estimated to be about 182K GPU hours. 

\paragraph{Model Evaluation.}
For the model evaluation phase, we use the super-net to initialize our child architectures, and then pre-train our model for 40,000 steps. We use the vanilla Masked Language Model (MLM) and Next Sentence Prediction (NSP) as our pre-training tasks. For fast pre-training, we set learning rate to 2e-4, batch size to 160, and warm-up proportion to 0.02. The learning rate decay strategy is set as linear. The fast pre-trained model (after 40,000 steps) will then be evaluated by the proxy task(GLUE) after finetuning 3 epochs. For fast finetuning, we set weight decay to 0.01, warm-up proportion to 0.01, and learning rate to 2e-5. The learning rate decay strategy is set as linear.

\subsection{Details in Fully-train Phase}
For pre-training configurations, we mostly use the same hyper-parameters as BERT. See Table~\ref{tab:pretrain_implementation_details} for more details. We use MLM and NSP to pre-train the searched architectures for 40 epochs (same with BERT). We set input sequence length as 128 and use Adam\citep{DBLP:adam} optimizer. For small-size, base-size and large-size models, the detailed configurations are listed in Table~\ref{tab:pretrain_implementation_details}. For large model(24 layers), we treat each two continuous layers (a convolution layer followed by a attention layer)  of the AutoBERT-Zero-base(12 layers) as a block and expand the base model to large model by inserting the same block after the original one.
 In the finetuning phase, following \citep{devlin2019bert, jiang2020convbert}, we search for learning rate among \{2e-5, 1e-5, 1.5e-5, 3e-5, 4e-5, 5e-5\} and weight decay among \{0.1, 0.01\}. Follow \cite{devlin2019bert,jiang2020convbert}, we finetune GLUE for 3 epochs. For SQuAD, in contrast to \cite{lan2019albert} and \cite{devlin2019bert} which use extra augmentation data, we finetuned on the original SQuAD v1.1 data for 3 epochs with batch size of 32. For SQuAD v2.0, we finetuned 2 epochs with batch size of 48. Following \citep{liu2019roberta}, we finetune RTE, STS-B, and MRPC using a MNLI checkpoint. For the evaluation interval, we use from \{100, 10\}.  We use the WordPiece embedding \citep{wu2016google}, and $30,000$ tokens are
contained in the dictionary. The special token {\tt [CLS]} is used as the first
token of each sequence. Another special token {\tt [SEP]} is used to separate
sentences in a sequence. The input token representation is the sum of token embedding, segmentation embedding and position embedding.

\begin{table}[h]
\begin{center}
\centering{}
\resizebox{0.45\textwidth}{!}{
\begin{tabular}{lccccc}
\hline
Method & FLOPs & Latency & Memory & \#Params & GLUE \tabularnewline
\hline
BERT-base    & 2.9e10 & 29.3s  & 2.5G  & 110M & 83.1  \tabularnewline
AutoBERT-Zero  & 2.3e10 & 27.2s & 2.1G  & 104M & 86.3 \tabularnewline
\hline
\end{tabular}}
\caption{{More comparison about model latency and memory. GLUE is reported on dev set. }}\label{tab:latency}
\end{center}
\end{table}

\subsection{Latency and Memory}
Table~\ref{tab:latency}  measure models with FLOPs, parameter number, latency and memory on Nvidia V100. For FLOPs, we follow the setting of \cite{clark2019electra,jiang2020convbert} and counts the inference FLOPs.  For latency,  we follow DynaBERT\cite{DBLP:conf/NeurIPS/HouHSJCL20} and do experiments on QNLI training set with batch size 128. From the results, we can find that our model AutoBERT-Zero is faster and occupies less memory during inference.

\bibliography{aaai22}

\end{document}